\documentclass[sigconf, nonacm]{acmart}
\usepackage{tabularx}
\usepackage{balance}
\usepackage{multirow}
\usepackage{adjustbox}

\usepackage{amsmath,bm}
\usepackage[english]{babel}
\usepackage{microtype}
\usepackage{enumitem}
\usepackage[noend,lined,boxed,vlined,ruled,linesnumbered]
{algorithm2e}
\usepackage{algpseudocode}
\usepackage{tabulary}
\usepackage{booktabs}
\usepackage{graphicx}
\usepackage{changepage}
\newcommand\vldbdoi{XX.XX/XXX.XX}
\newcommand\vldbpages{XXX-XXX}
\newcommand\vldbvolume{14}
\newcommand\vldbissue{1}
\newcommand\vldbyear{2025}
\newcommand\vldbauthors{\authors}
\newcommand\vldbtitle{\shorttitle} 
\newcommand\vldbavailabilityurl{URL_TO_YOUR_ARTIFACTS}
\newcommand\vldbpagestyle{plain} 
\usepackage[labelfont=normalfont,textfont=normalfont]{subcaption}
\usepackage{listings}
\usepackage{xcolor}
\usepackage{float}

\newcommand{\name}[1]{\textsf{MontePrep}}

\begin{document}
\title{MontePrep: Monte-Carlo-Driven Automatic Data Preparation without Target Data Instances}

\author{Congcong Ge}
\affiliation{%
  \institution{Zhejiang University}
}
\email{gcc@zju.edu.cn}

\author{Yachuan Liu}
\affiliation{%
  \institution{Zhejiang University}
}
\email{liuyachuan@zju.edu.cn}

\author{Yixuan Tang}
\affiliation{%
  \institution{National University of Singapore}
}
\email{yixuan@comp.nus.edu.sg}

\author{Yifan Zhu}
\affiliation{%
  \institution{Zhejiang University}
}
\email{xtf_z@zju.edu.cn}

\author{Yaofeng Tu}
\affiliation{%
  \institution{ZTE Corporation}
}
\email{tu.yaofeng@zte.com.cn}

\author{Yunjun Gao}
\affiliation{%
  \institution{Zhejiang University}
}
\email{gaoyj@zju.edu.cn}





\begin{abstract}

In commercial systems, a pervasive requirement for automatic data preparation (ADP) is to transfer relational data from disparate sources to targets with standardized schema specifications.
Previous methods rely on labor-intensive supervision signals or target table data access permissions, limiting their usage in real-world scenarios.

To tackle these challenges, we propose an effective end-to-end ADP framework \name~, which enables training-free pipeline synthesis with zero target-instance requirements.
\name~ is formulated as an open-source large language model (LLM) powered tree-structured search problem.
It consists of three pivot components, i.e., a data preparation action sandbox (\textsf{DPAS}), a fundamental pipeline generator (\textsf{FPG}), and an execution-aware pipeline optimizer (\textsf{EPO}).
We first introduce \textsf{DPAS}, a lightweight action sandbox, to navigate the search-based pipeline generation.
The design of \textsf{DPAS} circumvents exploration of infeasible pipelines.
Then, we present \textsf{FPG} to build executable DP pipelines incrementally, which explores the predefined action sandbox by the LLM-powered Monte Carlo Tree Search.
Furthermore, we propose \textsf{EPO}, which invokes pipeline execution results from sources to targets to evaluate the reliability of the generated pipelines in \textsf{FPG}.
In this way, unreasonable pipelines are eliminated, thus facilitating the search process from both efficiency and effectiveness perspectives.
Extensive experimental results demonstrate the superiority of \name~ with significant improvement against five state-of-the-art competitors. 
\end{abstract}

\maketitle

\pagestyle{\vldbpagestyle}
\begingroup\small\noindent\raggedright\textbf{Reference Format:}\\
\vldbauthors. \vldbtitle. XXX, \vldbvolume(\vldbissue): \vldbpages, \vldbyear.\\
\href{https://doi.org/\vldbdoi}{doi:\vldbdoi}
\endgroup
\begingroup
\renewcommand\thefootnote{}\footnote{\noindent
Preprint. \\
}\addtocounter{footnote}{-1}\endgroup

\ifdefempty{\vldbavailabilityurl}{}{
\vspace{.1cm}
\begingroup\small\noindent\raggedright\textbf{ Artifact Availability:}\\
The source code, data, and/or other artifacts have been made available at \url{https://github.com/ZJU-DAILY/MontePrep}.
\endgroup
}

\begin{figure}[H]
  \centering
  \includegraphics[width=\linewidth]{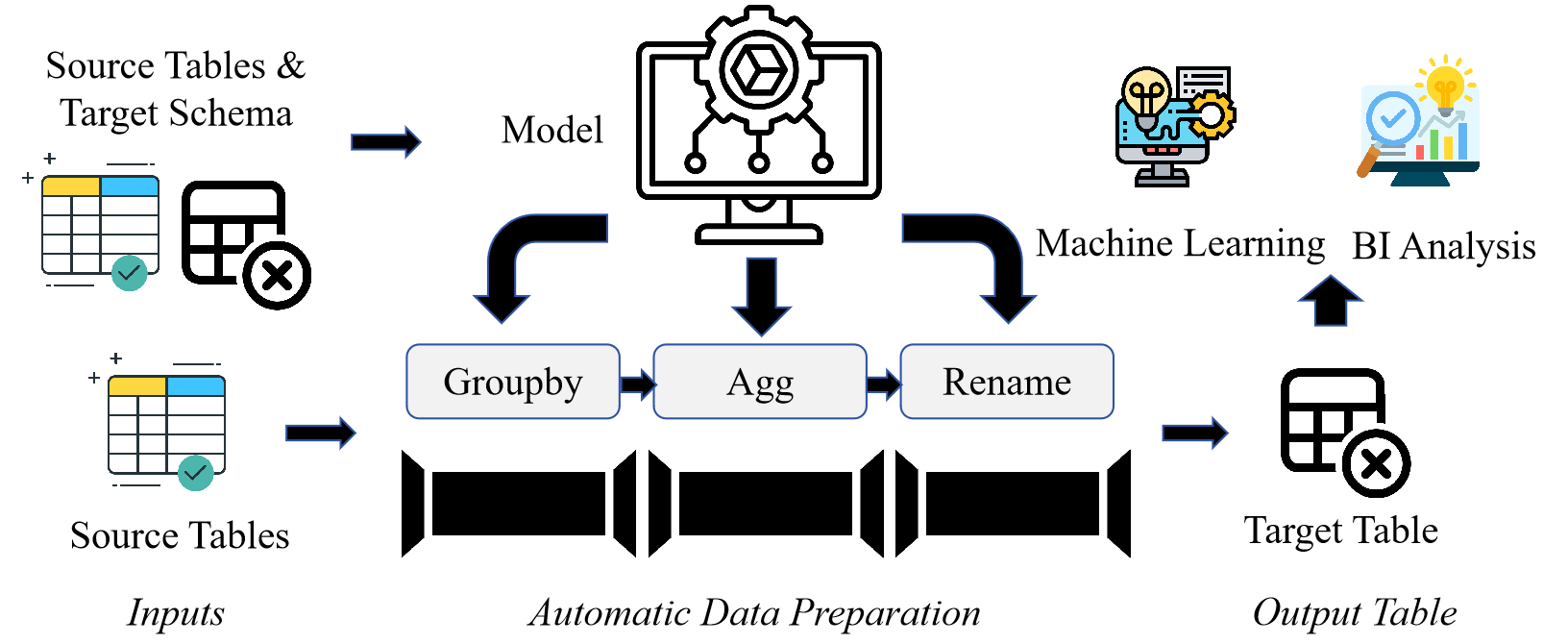}
  \vspace{-6mm}
  \caption{Illustration of the ADP task that transforms data from disparate sources into targets with standardized schema specifications while requiring \emph{zero} target instance access.}
  \label{Scenario}
  \vspace{-2mm}
\end{figure}

\section{Introduction}
\label{sec:intro}
Self-service data preparation (DP)~\cite{gartner2016market, deng2017data}, a prerequisite for data democratization, is becoming increasingly important. It aims to allow data scientists to prepare data without extensive domain knowledge or time-consuming efforts.
\emph{Automatic data preparation} (ADP)~\cite{autopipeline21, autosuggest20} is an effective solution to achieve the goal of self-service data preparation.
It transfers raw data from disparate sources into formats that are ready for business intelligence (BI) analysis or machine learning purposes by generating sequential pipelines, which are composed of transformation operators (e.g., Join, Groupby, and Rename).
Recently, various ADP techniques have emerged, such as transform-by-example~\cite{TDE18}, transform-by-pattern~\cite{jin2020auto}, transform-by-target~\cite{autopipeline21}, transform‑for‑joins~\cite{autojoin17}, and LLM-based transformation~\cite{sharma2023automatic}.
Nonetheless, there is still a big room for implementing ADP tools in practice due to \emph{unstandardized target schema specifications} and \emph{restricted data access privileges}.



\textbf{New ADP task - complies with standardized target schema without access to target instances.}
In practical commercial systems, a pervasive requirement for ADP is transforming source data into target tables conforming to standardized schema specifications.
We emphasize that standardized schemas serve as the structural backbone of the enterprise data flow. It ensures the compatibility of the cross-system interaction.
Furthermore, enterprise data providers always retain access to source table instances but are denied access to target table instances, due to compliance controls, data residency laws, or internal policy restrictions.
This scenario commonly occurs when: (i) target tables contain sensitive information (e.g., personally identifiable information) or (ii) a data provider and a data receiver belong to different departments with inter-organizational data silos.
Consequently, we focus on the new ADP task that transforms data from disparate sources into targets with standardized schema specifications while requiring \emph{zero} target instance access.
Figure~\ref{Scenario} presents an overview of this new ADP task.

\begin{figure*}[t]
  \centering
  \vspace*{-1em}
  \includegraphics[width=\linewidth]{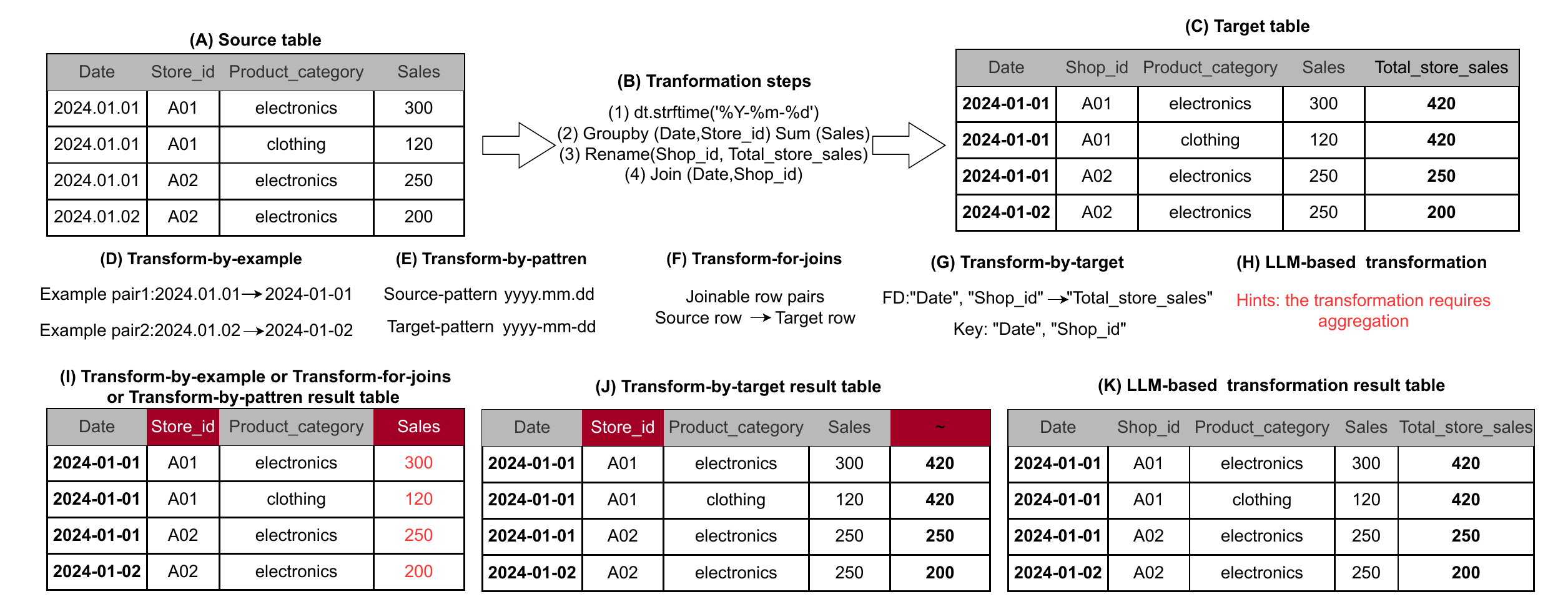}
  \vspace*{-2em}
  \caption{A variety of approaches can be useful for ADP: transform-by-example, transform-by-pattern, transform-by-target, transform-for-joins, and LLM-based approaches. Nonetheless, these approaches have non-negligible drawbacks when dealing with DP problems in data-sensitive or unsupervised situations.}
  \label{fig:retail_table_example}
  \vspace*{-1em}
\end{figure*}


\vspace{-1mm}
\begin{example}
\label{example:intro}
We consider a data preparation task in a supply chain scenario, where an ERP system integrates daily sales data from different regional stores into a central data warehouse.
In this scenario, the central data warehouse follows a standardized schema, which enables unified analytics and streamlined business management.
However, it is worth noting that data privacy constraints across regional stores prevent access to target table instances.
Figure~\ref{fig:retail_table_example} depicts an example.
Here, raw data migrates from a source table (Figure~\ref{fig:retail_table_example}(A)) to a standardized target table (Figure~\ref{fig:retail_table_example}(C)) via multiple transformation steps (Figure~\ref{fig:retail_table_example}(B)).
The source table $T_l$ contains records with four columns, i.e., \texttt{Date}, \texttt{Store\_id}, \texttt{Product\_category}, and \texttt{Sales}.
The target table $T_r$ is semantically enriched w.r.t. the source table: (i) it standardizes the \texttt{Date} format (from \texttt{yyyy.mm.dd} to \texttt{yyyy-mm-dd}); (ii) renames \texttt{Store\_id} to \texttt{Shop\_id}; and (iii) introduces a new column \texttt{Total\_store\_sales}, which aggregates daily \texttt{Sales} per store.
\end{example}
\vspace{-1mm}

We want to argue that automating such task in Example~\ref{example:intro} is non-trivial in practical scenarios.
Existing ADP approaches fall short due to the following challenges:

\vspace{1mm}
\noindent
\textbf{Challenge I:}
\textit{How to implement ADP without exposing target table data instances?}
Existing approaches predominantly rely on access to the target table to guide the ADP process.
It introduces significant privacy and compliance risks in practice.
Specifically, \emph{transform-by-example}~\cite{flashfill, becker2019microsoft, trifacta, TDE18, wang2017synthesizing} requires a set of source-target mapping pairs as examples to synthesize the logic of pipeline generation from $T_l$ to $T_r$.
Taking Figure~\ref{fig:retail_table_example}(D) as an example, pairs like mapping \texttt{2024.01.01} to \texttt{2024-01-01} are needed.
Similarly, \emph{transform-for-join}~\cite{autojoin17} requires joinable record pairs to infer the join logic between source tables and target tables, as shown in Figure~\ref{fig:retail_table_example}(F).
Figure~\ref{fig:retail_table_example}(I) shows the result table derived from the above two methods. It only achieves the right date format but fails to perform the correct operations, including Groupby, Rename, and Join.
Recently, large language model (LLM), as a powerful tool, has been utilized in the ADP task~\cite{sharma2023automatic}.
This LLM-based approach treats ADP as a SQL generation problem.
It first injects a specific pipeline generation description and several data hints into a prompt, and then sends the prompt to a specific closed-source LLM via an external API.
Finally, the API returns a SQL-scripts-based pipeline as a result. 
Taking Figure~\ref{fig:retail_table_example}(H) as an example,
this method provides a data hint that generating the target table $T_r$ requires an aggregation operation, which is crucial for navigating a meaningful pipeline generation result.
As shown in Figure~\ref{fig:retail_table_example}(K), it generates the correct output table.
However, providing additional data hints requires technical personnel involvement, which introduces additional complexity.
Moreover, transmitting such contextual information to external services violates internal compliance policies and data residency regulations, posing unacceptable risks to industrial systems.

\vspace{1mm}
\noindent
\textbf{Challenge II:} 
\textit{How to implement ADP without involving any supervised data?}
Many existing ADP methods rely on supervised data, such as labeled examples or auxiliary metadata, to guide transformation pipeline synthesis.
It limits their applicability in settings where such supervision is unavailable, incomplete, or expensive to acquire.
For example, \emph{transform-by-example}~\cite{flashfill, becker2019microsoft, trifacta, TDE18, wang2017synthesizing} and \emph{transform-for-join}~\cite{autojoin17} rely on supervised signals such as explicit source-target example pairs or joinable record pairs to infer transformation logic. As illustrated in Figure~\ref{fig:retail_table_example}(D) and (F),  these methods require instance-level correspondences to synthesize meaningful operations, which is costly in real-world settings.
Similarly, \emph{transform-by-pattern} \cite{jin2020auto} achieves ADP results by learning reusable transformation patterns, as shown in Figure~\ref{fig:retail_table_example}(E).
Nonetheless, acquiring sufficient and correct transformation patterns requires the support of a large corpus.
\emph{Transform-by-target}~\cite{autosuggest20, autopipeline21} consists of two steps. First, it relies on supervision to predict single operators, such as Join and Pivot~\cite{autosuggest20}. 
Then, it involves both the set of single operators obtained in the previous step and several integrity constraints (e.g., functional dependencies and keys) to generate the entire pipeline~\cite{autopipeline21}, as shown in Figure~\ref{fig:retail_table_example}(G). Thus, it still rely on supervision signals, limiting its applicability in real-world scenarios. 
Figure~\ref{fig:retail_table_example}(J) presents the resulting table.
In addition, since the renaming operator is not supported in this method, it is not able to generate standardized schemas of target tables, which requires additional mapping procedures.


To address these two challenges, an intuitive idea is to incorporate open-source LLMs into ADP techniques.
Specifically, open-source LLMs can be deployed locally, thereby eliminating data leakage risks from external API access.
Moreover, their robust semantic comprehension capabilities have proven effective across diverse tasks without requiring labor-intensive training.
To this end, we propose \name~ to automate the data preparation pipeline without any target table instance and training effort.

\name~ is formulated as an open-source LLM powered tree-structured search problem to effectively explore the space of DP transformation operators.
\name~ consists of three pivotal components: (i) an \emph{DP action sandbox} termed as \textsf{DPAS};
(ii) a \emph{fundamental pipeline generator} called \textsf{FPG}; and
(iii) an \emph{execution-aware pipeline optimizer} named \textsf{EPO};
First, \textsf{DPAS} is present to restrict the search scope of DP pipeline generation within a bounded decision space.
Second, we propose \textsf{FPG} to generate DP pipelines using a Monte Carlo tree search-based method.
Here, each node corresponds to a partial DP transformation state, and each edge represents a DP transformation action from the proposed ADP action sandbox.
At each search step, a large language model (LLM) analyzes the source table and the target table schema to propose the next transformation action. Once a complete DP pipeline is generated, the LLM evaluates the reliability of the generated pipeline using \emph{self-reward mechanism}.
Finally, \name~ selects the most reliable pipeline as the final result.
Furthermore, \textsf{EPO} is introduced to optimize the entire search process.
It contains two perspectives of optimization.
One optimizes the self-reward mechanism by invoking the pipeline execution results from sources to targets to evaluate the reliability of the generated pipelines.
In this way, unreasonable pipelines will be eliminated.
The other one is used to accelerate the search progress by a \emph{simulation cache} mechanism and an \emph{early termination} method. 
Our contributions are summarized as follows:


\begin{itemize}[nosep, topsep=0pt]
\item \emph{User-friendly Framework}. 
\name~ is a novel end-to-end framework that generates DP pipelines with \emph{zero} target instances and \emph{zero} training efforts.
The zero-target-instance requirement enables data providers to solve the DP problem in cross-domain scenarios without access to the target data, thus ensuring data confidentiality.
This training-free automation eliminates expertise barriers, enabling non-technical users to participate in data preparation.
\item \emph{Lightweight Action Sandbox}.
We define a lightweight DP action sandbox \textsf{DPAS}, which restricts the search scope of the DP pipeline generation. \textsf{DPAS} circumvents exploration of infeasible transformation operators.
\item \emph{Execution-aware Optimizer}.
We present \textsf{EPO}, an execution-aware optimizer, to facilitate the search progress from both efficiency and effectiveness perspectives.
\textsf{EPO} invokes the pipeline execution results from sources to targets to evaluate the reliability of the generated pipelines.
In this way, unreasonable pipelines will be eliminated.
\item \emph{Extensive Experiments}.
We conduct comprehensive experimental evaluation on ADP tasks against state-of-the-art approaches.
Extensive experimental results demonstrate the superiority of \name~.
\end{itemize}

\noindent
\textbf{Organization.} The rest of the paper is organized as follows.
Section~\ref{sec:preliminaries} presents the preliminaries.
Section~\ref{sec:framework} introduces the technical details of our proposed \name~ framework.
Section~\ref{sec:experiments} reports the experimental results
and our findings.
Section~\ref{sec:related_work} reviews the related work. Finally, Section~\ref{sec:conclusions} concludes the paper.

\section{preliminaries}
\label{sec:preliminaries}
In this section, we present some preliminaries, including problem statement, the supported DP transformation operators in \name~, and background techniques.

\subsection{Problem Statement}
As described in Section~\ref{sec:intro}, we focus on the ADP task that transforms data from disparate sources into targets with standardized schema specifications while requiring \emph{zero} target-instance access.
In commercial systems, despite the access limitation, business analysts and data engineers typically possess a clear understanding of the target schema that the enterprise data should conform.
To this end, it is realistic to assume that we have access to the schema of target tables even when the data instances are unavailable.

Formally, a relational table is denoted as $T = (S, D)$, where \( S = \{s_1, s_2, \ldots, s_m\} \) is the schema of $T$ and $D$ denotes the corresponding data rows, a.k.a., data instances.
Let $T_l = (S_l, D_l)$ be a source table with schema \( S_l = \{s_{l_1}, s_{l_2}, \ldots, s_{l_m}\} \).
and a sampled dataset $D^s_l \subseteq D_l$.
Let $T_r = (S_r, D_r)$ be a target table with schema \( S_r = \{s_{r_1}, s_{r_2}, \ldots, s_{r_m}\} \).
Recall that $D_r$ is excluded from consideration in our settings due to the lack of access to target data instances. Unless otherwise stated, the target tables $T_r$ mentioned below do not contain any data instances.
The objective is to synthesize a data preparation pipeline $\mathcal{P}=(o_1(p_1), o_2(p_2), \ldots,$ $o_k(p_k))$ from $T_l$ to $T_r$.
Here, $\mathcal{P}$ is an executable program composed of transformation operators. Each \( o_i \) is selected from a set of supported transformation operators (see Section~\ref{sec:operators}), denoted as $\mathcal{O}$ = \{ \texttt{Rename},~\texttt{Join}, \texttt{Union},~\texttt{Groupby},~\ldots\}.
Each \( p_i \) denotes the operator-specific parameters, such as join keys and aggregation columns.

Applying the pipeline \( \mathcal{P} \) to the given source table $T_l$ should produce an output table \( \hat{T}_r \) that satisfies the target schema \( S_r \). Formally,
\begin{align}
o_k(p_k) \circ \cdots \circ o_2(p_2) \circ o_1(p_1)(T_l) \rightarrow \hat{T}_r.
\end{align}

Ideally, $\hat{T}_r \equiv T_r$, where $\equiv$ means an equivalence relation.
It is important to emphasize that no sampled data from the target table $T_r$ is available at synthesis time in our settings.
The entire data preparation pipeline is required to be inferred solely from the source table and the target schema without any supervision, constituting a significant challenge for the ADP task.

\subsection{Supported Transformation Operators}
\label{sec:operators}
We consider two broad categories of transformation operators in \name~, i.e.,  \emph{table-level DP operators} and \emph{column-level DP operators}, as summarized in Table~\ref{tab:operators}.

\begin{table}[h]
\centering
\caption{Overview of supported transformation operators}
\vspace{-1em}
\label{tab:operators}
\renewcommand{\arraystretch}{1.15}
\begin{tabular}{p{0.28\linewidth} p{0.6\linewidth}}
\toprule
\textbf{Category} & \textbf{Transformation Operators} \\
\midrule
Table-level & Join, GroupBy, Pivot, Unpivot, Union, Add/Drop Columns, Rename \\
Column-level & Column Arithmetic, Date Formatting \\
\bottomrule
\end{tabular}
\end{table}

\noindent
\textbf{Table-level DP operators}.
In this category, operations can reshape the schema of the given table.

\begin{enumerate}[leftmargin=*]
\item \textit{Join}~\cite{op-join} integrates information across multiple tables based on a shared key column. It corresponds to \texttt{pandas.merge()}.

\item \textit{GroupBy}~\cite{op-groupby} partitions data into groups based on one or more key columns and computes aggregate functions—such as \texttt{sum}, \texttt{mean}—for each group, aligning with \texttt{pandas.groupby()}.

\item \textit{Pivot}~\cite{op-pivot} transforms long-format data into wide-format by turning unique values in a categorical column into new columns, as implemented in \texttt{pandas.pivot()}.

\item \textit{Unpivot}~\cite{op-unpivot} reverses a pivot operation by collapsing multiple columns into key-value pairs. It aligns with \texttt{pandas} \texttt{.melt()}.

\item \textit{Union}~\cite{op-union} combines multiple tables by appending rows with compatible schemas, typically along the vertical axis. This operation corresponds to \texttt{pandas.concat(..., axis=0)}.

\item \textit{Add/Drop Columns} adjusts the schema by introducing constant-valued columns or removing irrelevant ones. This is typically implemented via assignment or column selection in \texttt{pandas}.

\item \textit{Rename}~\cite{op-rename}, similar to \texttt{pandas.rename()}, standardizes column names for schema alignment or semantic clarity.
\end{enumerate}

\noindent
\textbf{Column-level DP operators}.
In this category, operators focus on manipulating the contents of individual columns.

\begin{enumerate}[leftmargin=*]
\item \textit{Column Arithmetic} applies arithmetic expressions across one or more columns to derive new values (e.g., \texttt{price * quantity}), using element-wise operations in \texttt{pandas}.

\item \textit{Date Formatting}~\cite{op-datetime} reformats or parses temporal values to enable downstream temporal analysis, such as converting strings to datetime objects or extracting date parts (e.g., year or month), via \texttt{pandas.to\_datetime()} and \texttt{dt} accessors.
\end{enumerate}

\subsection{Monte Carlo Tree Search}
\label{sec:mcts}
Monte Carlo Tree Search (MCTS)~\cite{coulom2006efficient} is a heuristic algorithm used for solving sequential decision-making problems, such as game AI like AlphaGo and AlphaZero~\cite{granter2017alphago}, that involve large and complex search spaces.
MCTS contains the following four phases:
\begin{itemize}[leftmargin=*]
  \item \textbf{Selection}: Starting from the root, the algorithm explores the tree by selecting child nodes based on a specific strategy that can balance both exploration and exploitation.
  Here, researchers often utilize \emph{Upper Confidence Bounds applied to Trees} (UCT) formula~\cite{kocsis2006bandit} as the balanced strategy.

  \item \textbf{Expansion}: Once a leaf node is reached, one or more feasible new child nodes, each of which corresponds to a potential future move, are added to the tree, unless the terminal state is reached.

  \item \textbf{Simulation}: A simulation method-often termed \emph{rollout}-is executed from the newly expanded node, following a default policy (e.g. random) until a terminal state is reached, thereby evaluating the newly expanded node's potential.

  \item \textbf{Backpropagation}: The simulation result, i.e., the reward, is propagated back to the root through the visited path, updating the statistical information of each visited node, including node visit counts and an estimated node value.
\end{itemize}

During the search process, MCTS iteratively selects, expands, simulates, and backpropagates reward signals to explore the most promising paths. This approach supports constrained exploration over large but non-deterministic search spaces with feedback.

\begin{figure*}[t]
  \centering
  \vspace{-2.5em}
  \includegraphics[width=\linewidth]{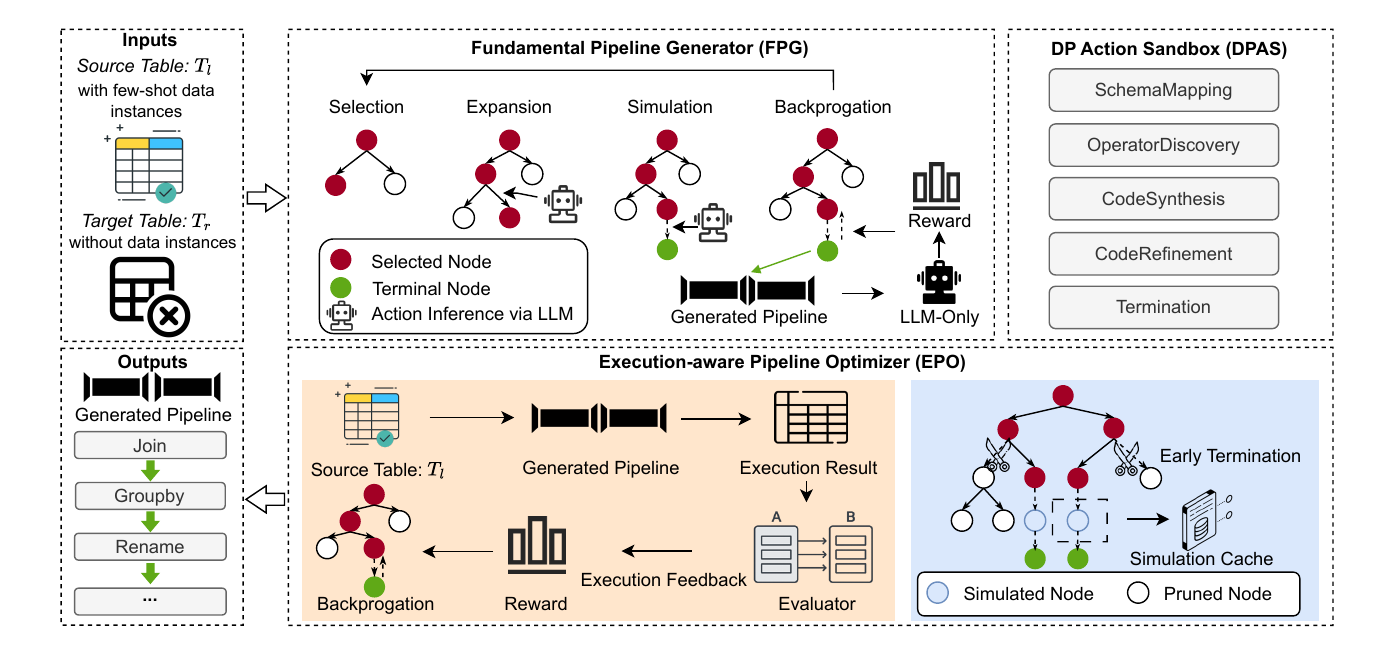}
  \vspace{-2.5em}
  \caption{Overview of the MontePrep framework for automatic data preparation.}
  \label{fig:framework}
\end{figure*}

\section{MontePrep Framework}
\label{sec:framework}
In this section, we describe our end-to-end ADP framework \name~ in detail.
Figure~\ref{fig:framework} illustrates the \name~ architecture, which includes three components: a \emph{DP action sandbox} (DPAS) followed by a \emph{fundamental pipeline generator} (FPG) and an \emph{execution-aware pipeline optimizer} (EPO).


\subsection{DP Action Sandbox}


Given that data preparation pipelines are composed of sequences of transformation operators, an intuitive method is to search all possible operator permutations, followed by the identification of the most plausible pipeline from the resultant combinatorial space.
Recent work \cite{autopipeline21} has demonstrated the empirical viability of search-oriented methods for ADP tasks.
It generates DP pipelines based on the set of transformation operators directly.
Motivated by the effectiveness of search-oriented methodologies, we define a discrete and symbolic sandbox (i.e., \textsf{DPAS}) that navigates and constrains the search process while elevating reasoning to a high level of abstraction.
In this sandbox, each action encapsulates a meaningful transformation sub-goal and operates in a reasoning state, allowing \name~ to build pipelines through explicit goal-oriented planning instead of ad-hoc code generation.

The sandbox is composed of five action types, termed as $\mathcal{A}=\{A_1, A_2, A_3, A_4, A_5\}$, each of which corresponds to a distinct transformation intent.
A specific action $a_i = (A^i_j, \theta_i)$ consists of an action type $A^i_j \in \mathcal{A}$ and a parameter object $\theta_i$ instantiated based on the current state. Next, we detail the supported action types.

\begin{itemize}[leftmargin=*]
\item {$A_1$: \texttt{SchemaMapping}.}  
In data preparation tasks, source and target table schemas often differ in naming, granularity, or structure. Automatically inferring how columns in the source table align with fields in the target schema is a prerequisite for meaningful transformation. This action establishes such correspondences by identifying semantically related columns. 

\item {$A_2$: \texttt{OperatorDiscovery}.}  
\name~ formulates transformation operator discovery as a semantic planning step.
This action leverages LLM to discover concrete transformation operators (as described in Section~\ref{sec:operators}) based on partial context, including the schema of the target table, column descriptions, and schema mapping results.
Here, the schema mapping results are optional. The reason is that the \texttt{OperatorDiscovery} action may be invoked either before or after the \texttt{SchemaMapping} action.
This flexibility reflects common strategies in human data wrangling, where one may either hypothesize the transformation logic upfront or derive it after examining the data.

\item {$A_3$: \texttt{CodeSynthesis}.}  
After identifying an appropriate transformation logic, \name~ generates executable Python code to implement it. This action instantiates the chosen operator using a specific syntax, i.e., \texttt{pandas} in our current implementation, based on the mapped schema and discovered logic.

\item {$A_4$: \texttt{CodeRefinement}.}  
In the \texttt{CodeSynthesis} action, the generated executable code may contain errors because of ambiguous schemas, inconsistent naming conventions, or incomplete specifications.
To address these issues, we propose the \texttt{CodeRefinement} action.
It aims to identify and correct such errors in the aforementioned executable code snippet.
This action is crucial for robustness, particularly in zero-shot or noisy settings, where initial code generations are often imperfect.

\item {$A_5$: \texttt{Termination}.}  
This action signals the end of the search process. It triggers a validation step to check whether the composed pipeline satisfies the schema of the target table. Specifically, if the output is incorrect or incomplete, MCTS will backtrack and explore alternative paths.
\end{itemize}
\vspace{-1.5em}


This abstraction enables modular construction of data preparation pipelines by structuring the synthesis process as a sequence of symbolic actions in \textsf{DPAS}.
Rather than treating pipeline generation as an atomic step (e.g., Join), we decompose the process into semantically meaningful sub-decisions, allowing each part to be controlled and revised during the pipeline synthesis process.


Furthermore, to ensure that action sequences in \name~ produce reasonable pipelines, we define a set of transformation constraints within the sandbox. These constraints are derived from structural patterns commonly observed in real-world data preparation workflows and specify reasonable transitions between action types.
For example, the \texttt{CodeRefinement} action must be executed after the \texttt{CodeSynthesis} action, and the \texttt{Termination} action should only be invoked after the pipeline has been synthesized or refined.
The complete transition constraints are summarized in Table~\ref{tab:action_constraints}.
Concretely, initial actions at the root node include \texttt{SchemaMapping}, \texttt{OperatorDiscovery}, and \texttt{CodeSynthesis}. 
Subsequent nodes are subject to transition constraints based on the current reasoning state and previously selected actions.
These constraints are dynamically enforced during node expansion to prune infeasible branches early in the search process of \name~.
By constraining which actions can logically follow others, we prevent inconsistent sequences and guide the \name~ toward constructing reasonable and executable DP pipelines.


\begin{table}[t]
\centering
\caption{Available next actions in the DP action sandbox.}
\vspace{-1em}
\label{tab:action_constraints}
\begin{tabularx}{\columnwidth}{lX}
\toprule
\textbf{Action Type} & \textbf{Valid Next Actions} \\
\midrule
--- & SchemaMapping, OperatorDiscovery, CodeSynthesis \\
SchemaMapping & OperatorDiscovery, CodeSynthesis \\
OperatorDiscovery & SchemaMapping, CodeSynthesis \\
CodeSynthesis & Termination, CodeRefinement \\
CodeRefinement & Termination \\
Termination & --- \\
\bottomrule
\end{tabularx}
\vspace{-2.5em}
\end{table}

\noindent
\textbf{Discussion.}
We emphasize that designing a bounded action space via \textsf{DPAS}, rather than directly using the action space composed of atomic transformation operators (atomic action space for short), significantly enhances the pipeline generation process in \name~, as to be detailed in Section~\ref{sec:fpg}.
Recall that \name~ is driven by large language models (LLMs) with a tree-structured search mechanism.
The search complexity of the tree-structured search method can be denoted as $O(|\mathbb{N}|^{|d|})$, where $|\mathbb{N}|$ is the number of nodes per level in the search tree and $|d|$ represents the depth of the entire tree structure.
Considering that data development engineers usually implement custom DP transformation operators through User-Defined Functions (UDFs), the atomic action space exhibits continuous expansion.
However, \textsf{DPAS} contains only five abstract operators. Such a constrained design drastically reduces the search complexity.
Moreover, our proposed transformation constraints, as illustrated in Table~\ref{tab:action_constraints}, can further reduce the search complexity.
In addition, the design of \textsf{DPAS} significantly reduces reasoning complexity for LLMs.
The reason is that selecting the correct operators from a small operator set is simpler for LLMs than choosing from a large collection. Furthermore, errors may propagate through operator sequences during the tree-structured search process, and thus fewer abstract operators contained in \textsf{DPAS} enables minimizing the risks of error propagation.




\subsection{Fundamental Pipeline Generator}

\label{sec:fpg}
\begin{figure}[t]
  \centering
  \includegraphics[width=\linewidth]{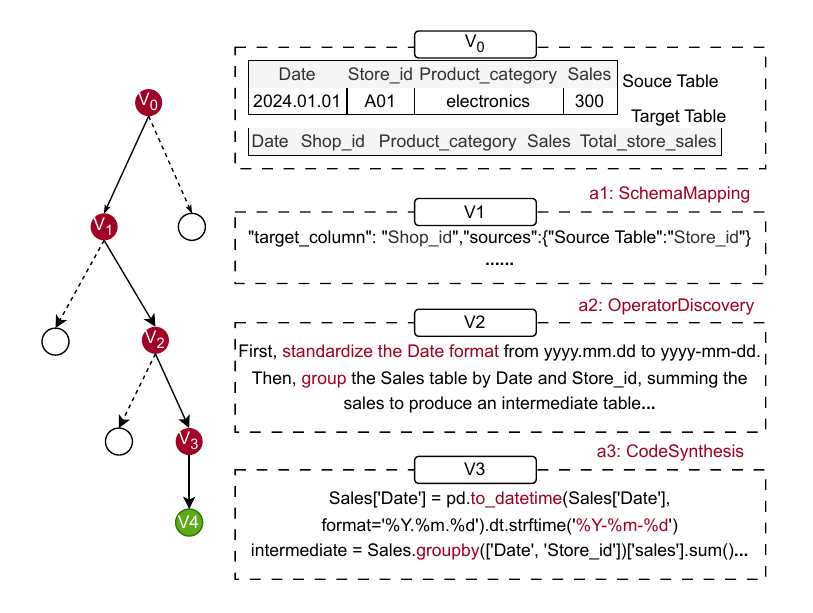}
  \vspace{-3em}
  \caption{An example of tree-structured search process.}
  \label{Action example}
  \vspace{-1.5em}
\end{figure}

Motivated by the effectiveness of MCTS in solving complex search problems (as illustrated in Section~\ref{sec:mcts}), we propose \emph{fundamental pipeline generator} (\textsf{FPG}), a Monte-Carlo-driven search approach for ADP, which orchestrates the MCTS process via open-source LLMs and executable feedback effectively.

We formulate the \textsf{FPG} process as a tree-structured search problem defined on a directed tree \( \mathcal{T} = (V, E) \), where each node \( v_i \in V \) represents a partial reasoning state and each edge \( e_i \in E \) means an action defined in \textsf{DPAS}.
A complete root-to-leaf path corresponds to a synthesized data preparation pipeline that produces a target-compliant output.
This tree-structured formulation enables \name~ to perform structured, step-by-step reasoning, starting from symbolic actions of \textsf{DPAS} and ending with concrete executable code.
Next, we introduce the settings of both nodes and edges of the search tree utilized in \textsf{FPG}.

\noindent(1) \textbf{Nodes (\( V \)) — Reasoning States.}  
Each node \( v_i \in V \) represents an intermediate transformation state, capturing the transformation logic constructed up to that node. As shown in Figure~\ref{Action example}, the root node encodes the initial state, consisting of the source table \( T_l \) and the target table \( T_r \). Intermediate nodes accumulate actions of \textsf{DPAS}, including inferred schema mappings, selected operators, etc. Leaf nodes represent completed data preparation pipeline expressed as executable logic (i.e., \texttt{pandas} in our current implementation).

\noindent(2) \textbf{Edges (\( E \)) — Reasoning Actions.}  
Each edge \( e_i \in E \) corresponds to an action decision of \textsf{DPAS}. These include actions such as selecting a schema mapping and choosing a data operator (e.g., \texttt{Groupby}, \texttt{Join}). Edges transition the state from one node to another and collectively determine a final candidate pipeline.

Figure~\ref{Action example} illustrates an example of ADP as a tree-structured search process. The source table contains four columns, termed as \texttt{Date}, \texttt{Store\_id}, \texttt{Product\_category}, and \texttt{Sales}. The target table includes a renamed column \texttt{Shop\_id} and a new column \texttt{Total\_stor\\e\_sales}, which aggregates daily sales totals for each store. The ADP process is modeled as a search tree, where:

\begin{itemize}[leftmargin=*]
  \item The root node \( v_0 \) represents the initial inputs, i.e., the source table with few-shot data instances and the target table containing only the schema without any data instances.
  \item The intermediate node \( v_1 \) performs the \texttt{SchemaMapping} action to establish correspondences between source and target columns. Next, Node \( v_2 \) applies the \texttt{OperatorDiscovery} action to identify a sequence of transformation operators, including operations such as \texttt{Date Formatting} and \texttt{GroupBy}. Due to space limitations, we omit subsequent inferred operators. Then, node \( v_3 \) utilizes the \texttt{CodeSynthesis} action to generate a data preparation pipeline.



  \item The final node \( v_4 \) denotes the termination. At this node, the generated pipeline is assigned a reward.
\end{itemize}

Thereafter, we are ready to introduce the details of the search process of \textsf{FPG}.
Algorithm~\ref{alg:adp_generator} presents its pseudo-code. 
It takes as inputs a source table $T_l$ with few-shot data instances, a target table $T_r$ containing only schema definitions, an LLM \( f_{\text{LLM}} \), and a rollout budget \( N \).
\textsf{FPG} performs $N$ rollouts to construct and evaluate candidate pipelines. Each rollout consists of the following four canonical phases: \emph{Selection}, \emph{Expansion}, \emph{Simulation}, and \emph{Backpropagation}.

\begin{algorithm}[t]
\caption{Fundamental Pipeline Generator (\textsf{FPG})}
\label{alg:adp_generator}
\KwIn{Source table $T_l$, target table $T_r$, LLM $f_{\text{LLM}}$, rollout budget $N$, and max depth $|d|$}
\KwOut{Set of valid transformation pipelines $\mathcal{P}_{\text{found}}$}

Initialize root node $v_0 \gets (T_l, T_r)$\;
Initialize empty pipeline set $\mathcal{P}_{\text{found}}$\;

\For{$iteration = 1$ \KwTo $N$}{
    $v_i \gets \textsc{Select}(v_0)$\;
    $\mathcal{C}(v_i) \gets \textsc{Expand}(v_i, f_{\text{LLM}})$\;
    $v'_{|d_m|} \gets \textsc{Simulate}(v_i, \mathcal{C}(v_i), f_{\text{LLM}})$\;
    $\text{reward} \gets \textsc{Evaluate}(v'_{|d_m|})$\;
    $\textsc{Backpropagate}(v'_{|d_m|}, \text{reward})$\;
    \If{pipeline is correct via evaluating reward}{
        $\mathcal{P}_{\text{found}} \gets \mathcal{P}_{\text{found}} \cup \textsc{GetPipeline}(v'_{|d_m|})$\;
    }
}
\Return{$\mathcal{P}_{\text{found}}$}

\end{algorithm}

\noindent
\textbf{Selection.}
The selection phase begins the MCTS rollout by traversing the current search tree from the root node \( v_0 \) toward a leaf or expandable node $v_i$ (Line 4 in Algorithm~\ref{alg:adp_generator}). At each node \( v \), the algorithm determines the next action \( a_i \in \mathcal{A} \) and chooses the child node $v'$ by maximizing an acquisition function based on the \emph{Upper Confidence Bounds applied to Trees} (UCT). Formally:
\begin{equation}
\text{UCT}(v, a) = Q(v_i, a_i) + c \cdot \sqrt{\frac{\log N(v_i)}{N(v_i, a_i)}}
\label{eq:ucb}
\end{equation}
Here, \( Q(v_i, a_i) \in \mathbb{R} \) is the cumulative reward for taking action \( a_i \) at node \( v_i \), where \( \mathbb{R}\) denotes the real number space. \( N(v_i) \) is the visit count of node \( v_i \), \( N(v_i, a_i) \) is the number of times action \( a_i \) has been applied at \( v_i \), and \( c \in \mathbb{R}^{+} \), where \( \mathbb{R}^{+}\) represents the set of positive real numbers, is a tunable constant that controls the trade-off between exploration and exploitation. Notably, if there exists any child node that has not been visited before (i.e., \( N(v_i, a_i) = 0 \)), the algorithm prioritizes selecting such nodes.

This selection strategy encourages the search to favor paths that have historically yielded high reward while also encouraging exploration of less-visited action branches. The logarithmic dependence on \( N(v_i) \) ensures diminishing optimism as more rollouts visit the same node, preventing over-exploration of suboptimal paths. The traversal continues recursively until an unexpanded node is reached. This node will be the frontier for subsequent expansion.


\begin{figure}[t]
  \centering
  \includegraphics[width=\linewidth]{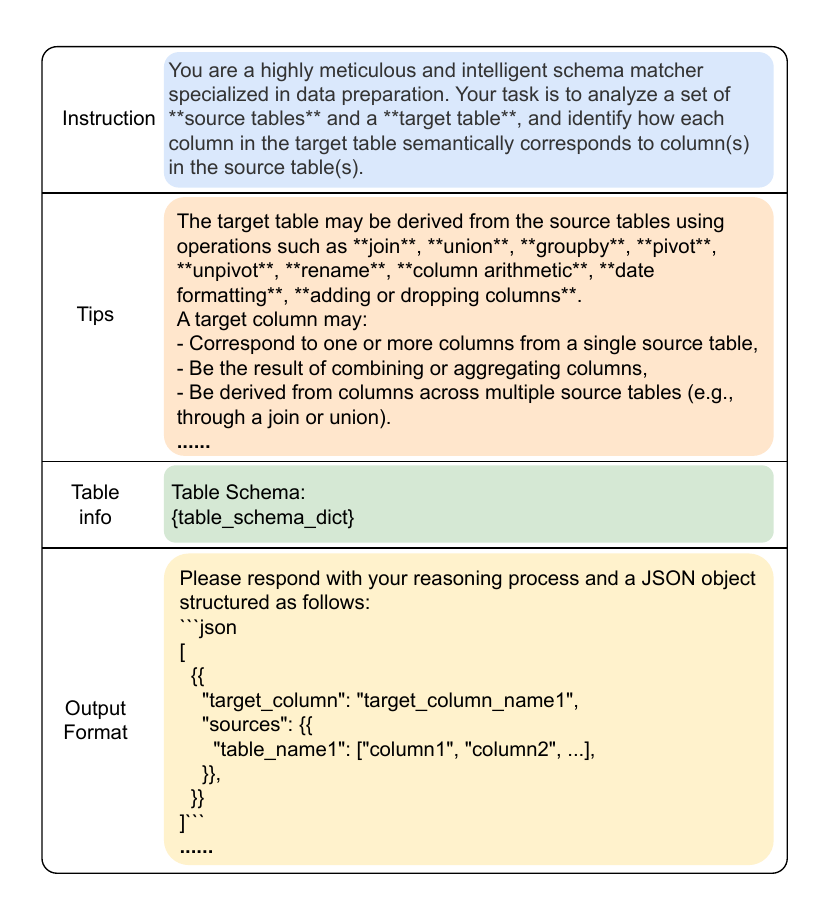}
  \vspace{-3.5em}
  \caption{An example of an action prompt template.}
  \label{Action prompt}
  \vspace{-2em}
\end{figure}

\noindent
\textbf{Expansion.}
Once the selection phase identifies a frontier node \( v_i \), the algorithm proceeds to the expansion phase (Line 5 in Algorithm~\ref{alg:adp_generator}). The goal of expansion is to enumerate and instantiate all valid next actions from \( v_i \).
This phase produces a candidate set of child nodes for the subsequent simulation phase.

Let \( \mathcal{A}_{\text{valid}}(v_i) \subseteq \mathcal{A} \) be the set of admissible action types at node \( v_i \), constrained by the transformation constraints (as detailed in Table~\ref{tab:action_constraints}). For each action type \( A^i_j \in \mathcal{A}_{\text{valid}}(v_i) \), the algorithm queries the LLM to generate its parameters:
\begin{equation}
\theta_i = f_{\text{LLM}}(\textsc{Prompt}(v_i, A^i_j) )
\end{equation}
where \( \theta_i \) is a structured object that specifies the action’s arguments (e.g., join keys, aggregate functions).
As illustrated in Figure~\ref{Action prompt}, we design a standardized action prompt template to facilitate this generation. Taking the \texttt{SchemaMapping} action as an example, the prompt begins with a detailed instruction that describes the task. Then, there are the carefully designed tips that guide the model in identifying semantic correspondences between source and target columns. This is followed by information about the table. To ensure the generated output is syntactically valid and aligned with downstream processing, the prompt concludes with an explicit specification of the expected response format.

Each parameterized action is defined as a tuple \( a_i = (A^i_j, \theta_i) \). Applying \( a_i \) to the current node produces a successor state \( v'_i = \delta(v_i, a_i) \), where \( \delta \) is the state transition process. All generated child nodes \( \{v'_1,~v'_2,~\dots\, v'_{|v|} \} \) are collected in a candidate set:
\begin{equation}\small
\mathcal{C}(v_i) = \left\{ \delta(v_i, a_i) \;\middle|\; a_i = (A^i_j, f_{\text{LLM}}(\textsc{Prompt}(v_i, A^i_j))),\; A^i_j \in \mathcal{A}_{\text{valid}}(v_i) \right\}
\end{equation}

The collected candidates \( \mathcal{C}(v_i) \) will be passed to the subsequent simulation phase for further evaluation.

\noindent
\textbf{Simulation.}
In the simulation phase (Line~6 in Algorithm~\ref{alg:adp_generator}), the algorithm randomly selects a node \( v'_i \) from the candidate set \(\mathcal{C}(v_i) \). Then it recursively applies actions to expand the selected node until a termination node is reached or the maximum rollout depth \( d_{\text{max}} \) is exceeded.
The result of this simulation is a sequence of nodes, which represents a candidate path as follows:
\begin{equation}
Path = \{v'_i, v'_{i+1}, \dots, v'_{|d_{m}|}\}, \quad \text{with} \quad v'_{i+1} = \delta(v'_i, a_i)
\end{equation}
Here, $|d_m|$ denotes the $m$-th depth of the search tree, and \( |d_m| \leq |d| \).
\( \delta \) is the state transition process as defined in the expansion stage.

At the end of the simulation phase, the terminal node \( v'_{|d_m|} \) typically contains a complete data preparation pipeline \( \mathcal{P} \). An intuitive approach for assessing the quality of such a pipeline is to take advantage of the reasoning capabilities of LLMs. To this end, we introduce a self-reward mechanism in which the LLM acts as a judge.
Specifically, given the source table \( T_l \), the target table \( T_r \), and the generated pipeline \( \mathcal{P} \), the LLM is prompted to evaluate whether \( \mathcal{P} \) is likely to produce a correct transformation. Based on its reasoning, the LLM returns a scalar reward \( r_{\text{LLM}} \in \{0, 0.5, 1.0\} \).
The evaluation criteria of the self-reward mechanism are as follows:
\begin{itemize}[leftmargin=*]
    \item \( r_{LLM} = 1~~~\) if the program is fully correct and complete;
    \item \( r_{LLM} = 0.5 \) if the program is partially correct but incomplete;
    \item \( r_{LLM} = 0~~~\) if the program is malformed or incorrect.
\end{itemize}

Designing an effective self-reward mechanism is challenging. Early experiments showed that prompts often led to unreliable evaluations: the model tended to assign high rewards to programs that were syntactically correct but semantically incorrect.

\begin{figure}[t]
  \centering
  \includegraphics[width=\linewidth]{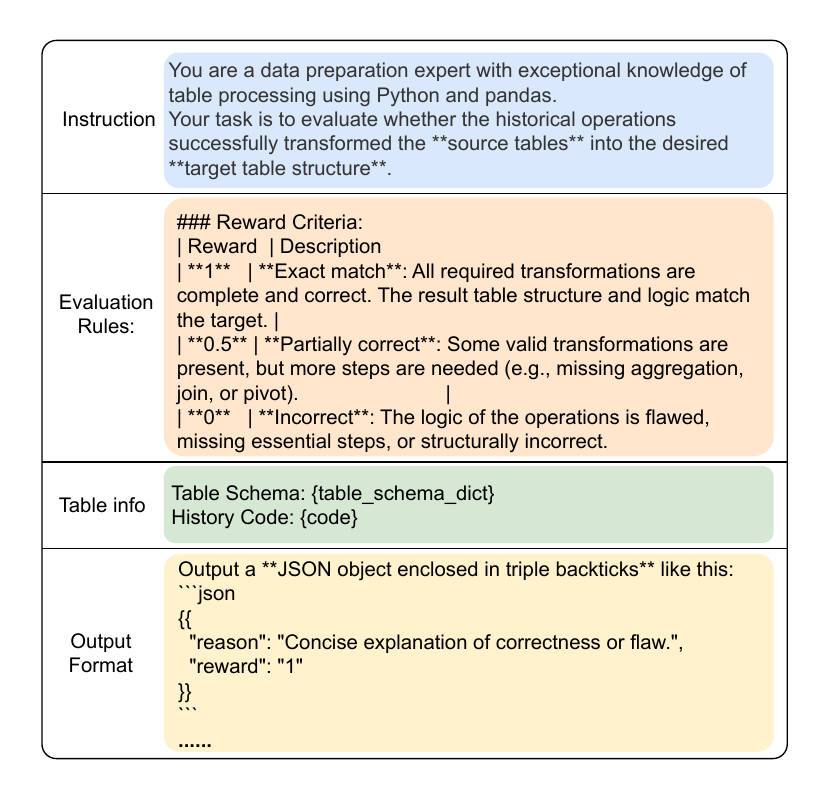}
  \vspace{-3em}
  \caption{Reward prompt template.}
  \label{Reward prompt}
\end{figure}

To address this, we design a clear, structured reward prompt composed of four main parts (see Figure~\ref{Reward prompt}). The first part, Instruction, defines the model’s role as a data preparation expert and clearly states its evaluation task. The second part, evaluation rules, introduces a clear reward rubric with three levels—exact match, partially correct, and incorrect—helping the model judge program quality. The third part, table information, provides the source table, target table schema, and generated pipeline. The final part, output format, specifies the expected structure of the model’s response. This structured prompt reduces ambiguity and solves the problems faced in the early experiment.

\noindent
\textbf{Backpropagation.}
Following the simulation phase, the final reward \( r_{LLM} \) assigned to the pipeline \( \mathcal{P} \) is propagated back along the search path. The goal of this phase (Line~7 in Algorithm~\ref{alg:adp_generator}) is to update value estimates and visitation counts for all intermediate node–action pairs encountered in the current rollout. These updates help refine the UCT-based search policy for future iterations.
Let the search path be denoted by a sequence of state–action pairs:
\begin{equation}
\Pi = \{(v'_0, a_0), (v'_1, a_1), \dots, (v'_{|d_m|}, a_{|d_m|})\}
\end{equation}

where each \( v'_i \in \mathcal{V} \) is a node, and \( a_i = (A_j^i, \theta_i)\) is the applied action at that node.
For each pair \( (v'_i, a_i) \in \Pi \), we maintain the following three statistics:
(i) \( Q(v'_i, a_i) \): the cumulative reward for taking action \( a_i \) at node \( v'_i \);
(ii) \( N(v'_i) \): the number of times node \( v'_i \) has been visited; and
(iii) \( N(v'_i, a_i)\): the number of times action \( a_i \) has been taken from \( v'_i \). We update these quantities as follows:
\begin{align}
Q(v'_i, a_i) &\leftarrow Q(v'_i, a_i) + r \label{eq:update_Q} \\
N(v'_i) &\leftarrow N(v'_i) + 1 \label{eq:update_Nv} \\
N(v'_i, a_i) &\leftarrow N(v'_i, a_i) + 1 \label{eq:update_Nva}
\end{align}

These updates ensure that nodes along promising paths—i.e., those associated with higher reward values—are more likely to be revisited in subsequent rollouts, while underperforming branches are naturally downweighted. The revised statistics directly impact the UCT score in Equation~\eqref{eq:ucb}, enabling adaptive exploration over the search space in ADP tasks.

\subsection{Execution-aware Pipeline Optimizer} 
\label{epo}






\begin{algorithm}[t]
\caption{Execution-aware Pipeline Optimizer (\textsf{EPO})}
\label{alg:epo}

\SetKwProg{Fn}{Function}{:}{}
\SetKwFunction{ExcutionAwareReward}{ExcutionAwareReward}
\SetKwFunction{Execute}{Execute}
\SetKwFunction{sim}{sim}
\Fn{\ExcutionAwareReward{$\mathcal{P}, T_l, T_r$}}{
    $\hat{T}_{r} \gets$ \Execute{$\mathcal{P}, T_l$}\;
    $\text{reward} \gets$ \sim{$\hat{T}_{r}, T_r$}\;
    \Return{$\text{reward}$}\;
}

\vspace{0.5em}

\SetKwFunction{SimulationCache}{SimulationCache}
\SetKwFunction{PROMPT}{PROMPT}
\Fn{\SimulationCache{$v_i, A_j^i, \mathcal{M}, f_{\text{LLM}}$}}{
    \If{$(v_i, A_j^i) \in \mathcal{M}$}{
        \Return{$\mathcal{M}[(v_i, A_j^i)]$}\;
    }
    $\theta_i \gets f_{\text{LLM}}($\PROMPT{$v_i, A_j^i$}$)$\;
    $\mathcal{M}[(v_i, A_j^i)] \gets \theta_i$\;
    \Return{$\theta_i$}\;
}

\vspace{0.5em}

\SetKwFunction{EarlyTermination}{EarlyTermination}
\Fn{\EarlyTermination{$\mathcal{P}, \mathcal{P}_{\text{found}}, K, T_l, T_r$}}{
    \If{pipeline is correct via evaluating reward}{
        $\mathcal{P}_{\text{found}} \gets \mathcal{P}_{\text{found}} \cup \mathcal{P}$\;
        \If{$|\mathcal{P}_{\text{found}}| \geq K$}{
            \Return{\texttt{True}}\;
        }
    }
    \Return{\texttt{False}}\;
}

\end{algorithm}

The fundamental pipeline generator builds data preparation pipelines and scores them using self reward. This approach treats the large language model as a self-evaluator: given the source table $T_l$, the target schema $S_r$, and the generated pipeline $\mathcal{P}$, the model infers a scalar reward that reflects its confidence in the pipeline's correctness. To further enhance the framework—particularly in scenarios where execution signals are available, we develop the \emph{Execution-aware Pipeline Optimizer} \textsf{(EPO)}. Specifically, \textsf{EPO} includes two key optimizations:
(i) execution-aware reward optimization, and  
(ii) search accelerator via simulation cache and early termination.
These enhancements are modular and apply during both the expansion and simulation phases of MCTS. A complete description of the procedure is provided in Algorithm~\ref{alg:epo}.

\noindent{\textbf{Execution-aware reward optimization.}}
Recall that the fundamental pipeline generator (\textsf{FPG}) involves a self reward to estimate the correctness of a data preparation pipeline $\mathcal{P}$.
This reward is solely based on LLM's reasoning ability.
It may give incorrect results due to hallucination issues.
This is especially common in multi-step pipelines, where small logical errors are easy to miss~\cite{huang2023large}.
Therefore, we incorporate execution feedback into the reward calculation, referring to it as the \emph{Execution-aware reward} (line~1 in Algorithm\ref{alg:epo}). As mentioned earlier, \name~ generates a data preparation pipeline \( \mathcal{P} \) based on the current search state. 
Once a complete pipeline is synthesized, it can be executed on the source table to produce an output table $\hat{T}_{r}$.
Formally,
\(\hat{T}_{r} = \textsc{Execute}(\mathcal{P}, T_l)\). We then compute the similarity between $\hat{T}_{r}$ and ${T}_{r}$ based on their schema similarity. Formally, $sim(\hat{T}_{r}, T_r) = \frac{|\hat{S}_{r} \cap S_r|}{| S_r|}$,
where \(\hat{S}_{r}\) denotes the set of column names in the output table $\hat{T}_{r}$, and \(S_r\) denotes the set of column names specified in the target table $T_r$.

This reward provides a grounded and deterministic signal that reflects whether the pipeline produces valid outputs.
As to be verified by our experimental study in Section~\ref{further-analysis}, we find that the execution-aware reward consistently outperforms self reward, especially under noisy or ambiguous conditions.

\noindent {\textbf{Search accelerator.}}
To improve the search efficiency of \textsf{FPG}, we introduce two strategies, namely \emph{simulation cache} (line~5 in Algorithm\ref{alg:epo}) and \emph{early termination} (line~11 in Algorithm\ref{alg:epo}), respectively.
The simulation cache maintains a parameter prediction cache \( \mathcal{M} \). It stores the result of each LLM call for a context \( (v_i, A_j^i) \in \mathcal{V} \times \mathcal{A} \).
In this way, when the same context is encountered in future rollouts, the cached parameter \( \theta_i \) is reused, circumventing redundant LLM inference calls. Formally:
\begin{equation}\small
  \theta_i = \mathcal{M}[(v_i, A^i_j)] \quad \text{if cached, else } \theta_i = f_{\text{LLM}}(\textsc{Prompt}(v_i, A^i_j))  
\end{equation}

Furthermore, to avoid unnecessary exploration once the search has produced sufficient successful candidates, \textsf{EPO} incorporates an early termination criterion. Let \( \mathcal{P}_{\text{found}} \) denote the set of pipelines with reward \( r = 1 \). If the number of such pipelines exceeds a user-defined threshold \( K \), the search process is halted. 

Together, simulation cache and early termination form a pair of synergistic strategies. They collectively reduce redundant exploration in automatic data preparation tasks, leading to a more efficient and targeted search process.

\noindent
\textbf{Discussion.}
Recall that we propose two reward strategies, i.e., self-reward and execution-aware reward.
Intuitively, another reward strategy is to combine the execution feedback and LLM together, referred to as a \emph{hybrid reward}.
The experimental results indicate that this hybrid reward performs better than self reward (see Section~\ref{further-analysis}).
Nonetheless, it is still not as good as the execution-aware reward. This is because the LLM’s reward judgments, even when informed by runtime results, cannot always guarantee correctness. Therefore, we use the execution-aware reward in the current implementation.
\section{Experiments}
\label{sec:experiments}







\noindent\textbf{Datasets.} We conduct experiments on two publicly available ADP benchmarks. The statistics are summarized in Table~\ref{tab:benchmarks}.

\begin{itemize}[leftmargin=*]
\item\textbf{Auto‑Pipeline Dataset}. This benchmark was introduced by \cite{autopipeline21}.
It comprises multi-step pipeline synthesis, with 700 transformation tasks mined from GitHub and industrial sources.
Each task includes a target schema specified in a `by-target' format, along with up to 10 transformation steps. To align the dataset with our problem statement, we performed several pre-processing steps.
First, we remove duplicate pipelines and all data instances from target tables.
Second, since this dataset primarily focused on data instances rather than schema clarity, several target schemas were represented with anonymous values (e.g., \texttt{col1}). To address this, we standardize target schemas by assigning meaningful column names. This normalization is common in the deployment of commercial system, such as master data management.

\item\textbf{Smart Building Dataset}.
This dataset is originally released in~\cite{sharma2023automatic}, aiming to build DP pipelines for energy data.
It comprises 105 real-world tasks involving complex structural transformations (such as aggregation, pivot, and column-wise computation). The target table in this dataset contains columns that are not related to the source table. This better reflects real-world scenarios, where target schemas often include extraneous or system-specific fields beyond the scope of available data.
Besides, the original benchmark includes detailed schema change hints—explicit natural language descriptions of how the target schema differs from the source tables, which serve as strong guidance for pipeline generation.
However, such hints are typically unavailable in real-world scenarios. Thus, we remove these hints in our evaluation and rely solely on the source table and target schema to guide the pipeline synthesis process.
\end{itemize}


\noindent\textbf{Evaluation Metrics.} To assess the correctness of synthesized pipelines, we use two evaluation metrics as follows:

\begin{itemize}[leftmargin=*]
\item \textbf{Execution Accuracy (EX).} We define EX as a strict indicator of end-to-end correctness. Given a generated pipeline $\mathcal{P}$, we execute $\mathcal{P}(T_l)$ to produce an output table $\hat{T}_r$, and compare it with the reference target table $T_r$. A task is marked as correct ($\text{EX} = 1$) only if $\hat{T}_r$ is identical to $T_r$ in both schema and data. This evaluation goes beyond schema alignment to ensure the correctness of the pipeline. Note that $T_r$ is used solely during evaluation; the generation process has no access to target instances.


\item \textbf{Column Similarity (CS).} Similar to the previous study~\cite{sharma2023automatic}, we define CS to measure schema-level alignment between the predicted output table $\hat{T}_r$ and the ground-truth target table $T_r$. A column is considered matched if its name exactly matches a column in the target schema. This metric is particularly useful for identifying partially correct pipelines. For example, a pipeline may fail to perform a group-by aggregation but still recover correctly named columns such as \texttt{Date} or \texttt{Store\_id}. In such cases, CS highlights a possible meaningful pipeline even when the full output is incorrect (i.e., EX = 0).

\end{itemize}


\begin{table}[t]
\centering
    \caption{Statistics of the datasets used in experiments.}
    \vspace{-3mm}
    \label{tab:benchmarks}
\begin{small}
\begin{tabular}{lcccc}
    \toprule
       Benchmark & \# of pipelines & \begin{tabular}{@{}c@{}}avg. \# of \\ input files \end{tabular}    & \begin{tabular}{@{}c@{}}avg. \# of \\ input cols \end{tabular} & \begin{tabular}{@{}c@{}}avg. \# of \\ input rows \end{tabular}  \\
        \midrule
        Auto-Pipline  & 481     & 3.4 & 8.3 &   8446.4     \\
        \midrule
        Smart Building   & 105     & 1 & 4.9 & 28.1     \\
    \bottomrule
    \end{tabular}
\end{small}
\vspace{-1em}
\end{table}

\noindent\textbf{Implementation Details.} 
We detail the hyper-parameters used in \name~ as follows.
Considering that the open-source Qwen-Coder family of large language models is widely used in both industrial and academia, we adopt the Qwen-Coder family (i.e., Qwen-Coder-7B, Qwen-Coder-14B, and Qwen-Coder-32B) for pipeline generation in the current implementation.
In the fundamental pipeline generator (\textsf{FPG}), we configure MCTS with the rollout budget $N = 10$, the maximum search depth $|d| = 5$, and the exploration constant of $c = 1$.
In the execution-aware pipeline optimizer (\textsf{EPO}), the early termination threshold is set to $K = 2$, terminating the search once two correct pipelines are fully discovered. We adopt the execution-aware reward strategy to evaluate pipeline correctness during runtime.
Unless explicitly specified, all hyper-parameters are set to their default values.
In addition, though the \name~ framework is designed to be local deployment, due to limitations in our experimental environment, we use the Qwen models via API calls. Importantly, all models used are open-source and self-hostable, ensuring that our implementation is consistent with the setting mentioned before.
All experiments are conducted on a Linux Server with an Intel(R) Xeon(R) Silver 4310 CPU (2.10GHz) and 128GB RAM.
The programs were all implemented in Python.


\begin{table*}[t]
\centering
\caption{Overall ADP results on Auto-Pipeline and Smart Building benchmarks.}
\label{tab:main-results}
\vspace{-1em}
\begin{tabularx}{\textwidth}{l l *{3}{>{\centering\arraybackslash}X} *{3}{>{\centering\arraybackslash}X}}
\toprule
\textbf{Method} & \textbf{LLM Model} & \multicolumn{3}{c}{\textbf{Auto-Pipeline Dataset}} & \multicolumn{3}{c}{\textbf{Smart Buildings Dataset}} \\
\cmidrule(lr){3-5} \cmidrule(lr){6-8}
& & EX (\%) & CS (\%) & Time (min) & EX (\%) & CS (\%) & Time (min) \\
\midrule
SQLMorpher & Qwen2.5-Coder-32B & 71.46 & 75.22 & 0.50 & \underline{55.45} & 58.54 & 0.42 \\
Chain-of-Table & Qwen2.5-Coder-32B & 56.13 & 58.77 & 0.12 & 43.81 & 48.03 & \textbf{0.16} \\
Chain-of-Thought & Qwen2.5-Coder-32B & 77.03 & 79.97 & \textbf{0.09} & 51.43 & 56.55 & \underline{0.20} \\
ReAct & Qwen2.5-Coder-32B & 79.83 & 84.40 & \underline{0.10} & 51.43 & 60.47 & 0.22 \\
FunctionCalling & Qwen2.5-Coder-32B & 18.27 & 50.29 & 0.28 & 19.61 & 30.08 & 0.40 \\
\midrule
\name~ & Qwen2.5-Coder-32B & \textbf{88.57} & \textbf{89.69} & 1.16 & \textbf{68.57} & \textbf{84.67} & 3.24 \\
\name~ & Qwen2.5-Coder-14B & \underline{82.95} & \underline{85.74} & 1.20 & 48.57 & \underline{60.84} & 3.17 \\
\name~ & Qwen2.5-Coder-7B  & 47.19 & 49.80 & 1.47 & 17.14 & 35.19 & 4.34 \\
\bottomrule
\end{tabularx}
\vspace{1em}
\begin{minipage}{\textwidth}
\footnotesize
\begin{adjustwidth}{1em}{0em}
\textsuperscript{1} We report Execution Accuracy (EX↑), Column Similarity (CS↑), and average pipeline prediction time (Time ↓, in minutes). 

\textsuperscript{2} \textbf{Bold} indicates current optimal values, while \underline{underscored} values are suboptimal.
\end{adjustwidth}
\end{minipage}
\vspace{-3em}
\end{table*}

\noindent\textbf{Methods Compared.} We compare \name~ against five representative methods as follows:
\begin{itemize}[leftmargin=*]
\label{baseline}
  \item \textbf{SQLMorpher}~\cite{sharma2023automatic}. An end-to-end LLM-driven system designed for building energy pipelines. SQLMorpher constructs task-specific prompts, optionally incorporating schema change hints, and iteratively refines generated SQL based on execution feedback. In our evaluation, we \textit{do not provide schema change hints} to the model, aligning with our goal of minimizing human intervention. This setting ensures a fair comparison with other methods that operate purely on the source table and target schema, as schema change hints require users to provide extra guidance, which reduces the level of automation.

  \item \textbf{Chain‑of‑Table}~\cite{wang2024chain}. Originally proposed for table understanding, Chain-of-Table decomposes complex transformations into a sequence of subtasks, aligning naturally with the multi-step structure of ADP. Rather than generating an entire program at once, it iteratively updates an intermediate table by applying one transformation at a time. To adapt this method for ADP, we replace its original operators (e.g., \textit{add\_col}, \textit{select\_row}) with ADP operators such as \texttt{GroupBy}, enabling pipeline construction through step-wise table state transitions.

  \item \textbf{Chain‑of‑Thought (CoT)}~\cite{wei2022chain}. CoT is a prompting framework that encourages step-by-step natural language reasoning before generating code. Although originally applied in arithmetic and logic-heavy domains, its structured thinking process is well-suited for ADP. CoT can be extended to ADP tasks by embedding operator descriptions (e.g., how \texttt{Pivot} function) into the prompt. This enables the LLM to plan transformations semantically before synthesizing corresponding code segments, supporting multi-step ADP scenarios with clear intermediate reasoning.

  \item \textbf{ReAct}~\cite{yao2023react}. ReAct integrates reasoning and acting by interleaving plans with actions and observations. In the context of ADP, this approach allows the LLM to reason over the execution results (e.g., output tables produced by operations such as \texttt{GroupBy} or \texttt{Join}) and refine its generated pipeline based on those results. By incorporating inspection of the transformation results, ReAct improves execution correctness in ADP pipelines.
  
  \item \textbf{FunctionCalling}~\cite{qin2024tool}. FunctionCalling constrains the LLM's action space by exposing a predefined set of transformation functions via structured JSON. Although originally developed for general tool-augmented tasks, its structured and modular design makes it an appropriate method for ADP, where transformations can be expressed as well-typed function calls. We adapt it to ADP by replacing the original function library with task-specific operators such as \texttt{GroupBy}, \texttt{Pivot}, and \texttt{Join}. This enables the model to construct pipelines in a controllable manner.
\end{itemize}

\begin{figure*}[t]
  \centering
  \includegraphics[width=\linewidth]{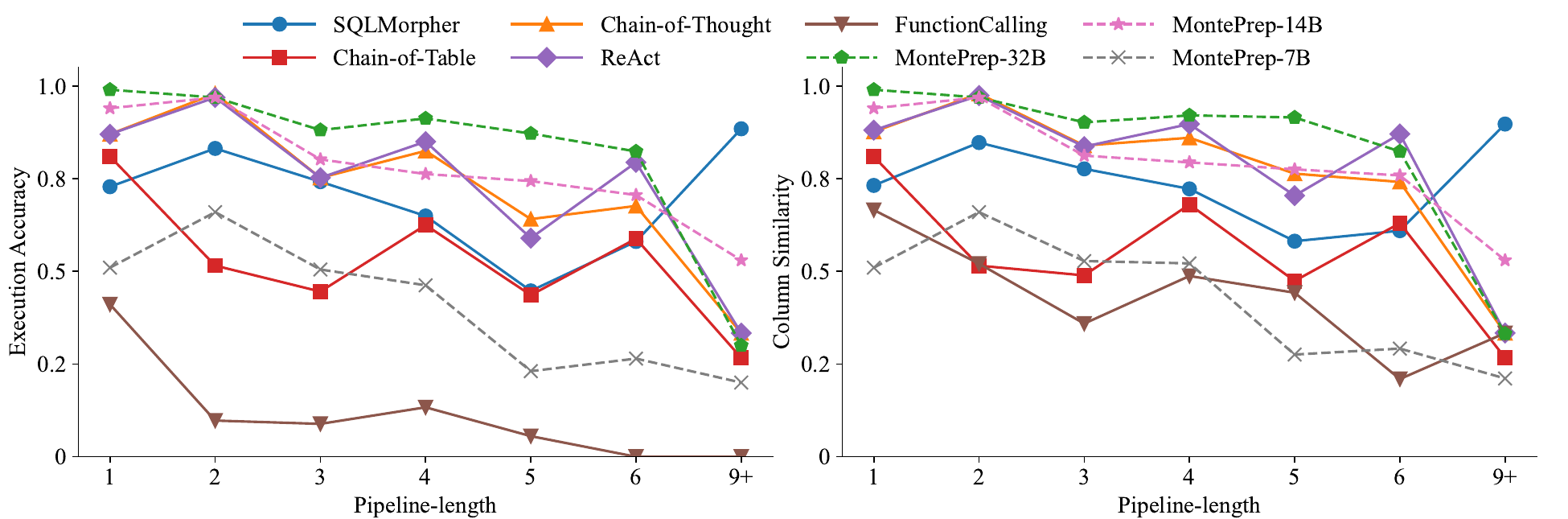}
  \vspace{-2em}
  \caption{Execution accuracy and column similarity grouped by pipeline length on the Auto-Pipeline dataset.}
  \label{fig:result_length}
\end{figure*}

\subsection{Main Results}

Table~\ref{tab:main-results} summarizes the overall experimental results on both the Auto-Pipeline and Smart Building benchmarks. We report the execution accuracy (EX), column similarity (CS), and average prediction time per task (time(min)). Below, we analyze these results in terms of effectiveness and efficiency. 

\noindent\textbf{Effectiveness.} Several important observations can be made as follows. (i) \name~ consistently outperforms all baseline methods in both execution accuracy and column similarity. For example, it achieves 88.57\% EX and 89.69\% CS on the Auto-Pipeline dataset, outperforming the strongest baseline (ReAct) by margins of 8.74\% and 5.29\%, respectively. This demonstrates that MontePrep’s structured search process, coupled with execution-aware optimization, enables more reliable generation of correct data preparation pipelines. (ii) The performance of MontePrep scales with model size. Notably, MontePrep-14B performs worse than MontePrep-32B but significantly better than MontePrep-7B. This indicates that larger models provide stronger reasoning capability in ADP task. In particular, MontePrep-14B achieves 82.95\% EX and 85.74\% CS on the Auto-Pipeline dataset. It outperforms all baseline methods using larger 32B LLMs. On the Smart Building dataset, MontePrep-14B also delivers competitive performance, with higher CS than other methods. These results show that MontePrep can significantly enhance the capabilities of LLM in ADP tasks, even at smaller scales. (iii) Among the baselines, ReAct shows strong performance, particularly on the Auto-Pipeline dataset. Its ability to interleave reasoning and acting contributes to improved effectiveness in ADP tasks. (iv) FunctionCalling performs poorly across both benchmarks, especially on Smart Building (EX = 19.61\%). This result suggests that a function invocation scheme is insufficient for complex multi-step reasoning tasks. (v) All methods, including MontePrep, perform better on Auto-Pipeline than on Smart Building. This is likely because the Smart Building tasks are based on real-world energy datasets. These datasets often contain ambiguous and irregular schema structures. In addition, they also include schema noise in the target table. Such characteristics increase the difficulty of identifying correct pipelines. All methods struggle in this setting, especially when no external schema change hints are provided.

Since the Auto-Pipeline benchmark groups tasks by pipeline length, we further analyze model performance across different levels of pipeline length. Figure~\ref{fig:result_length} present execution accuracy and column similarity for each pipeline length. MontePrep-32B maintains strong performance across most lengths, with a decline at length 9+. In contrast, SQLMorpher performs relatively well on length 9+ tasks, likely due to its strength in handling union operations, which are more common in longer pipelines. However, it performs poorly on shorter pipelines that require more precise and diverse transformations such as GroupBy, Rename, and Pivot. These results suggest that SQLMorpher performs well in specific scenarios but lacks general versatility. Overall, MontePrep demonstrates more stable and robust performance across varying levels of length. Note that the Smart Building dataset is not grouped by length and is therefore excluded from this analysis.

\noindent\textbf{Efficiency.} In terms of runtime, MontePrep exhibits a clear trade-off between accuracy and prediction time. Its iterative Monte Carlo Tree Search and pipeline execution introduce additional computational cost. MontePrep requires several minutes to complete a single task. In contrast, baseline methods often finish in under one minute—but at the cost of lower accuracy.

\subsection{Ablation study}
\begin{figure}[t]
  \centering
  \includegraphics[width=\linewidth]{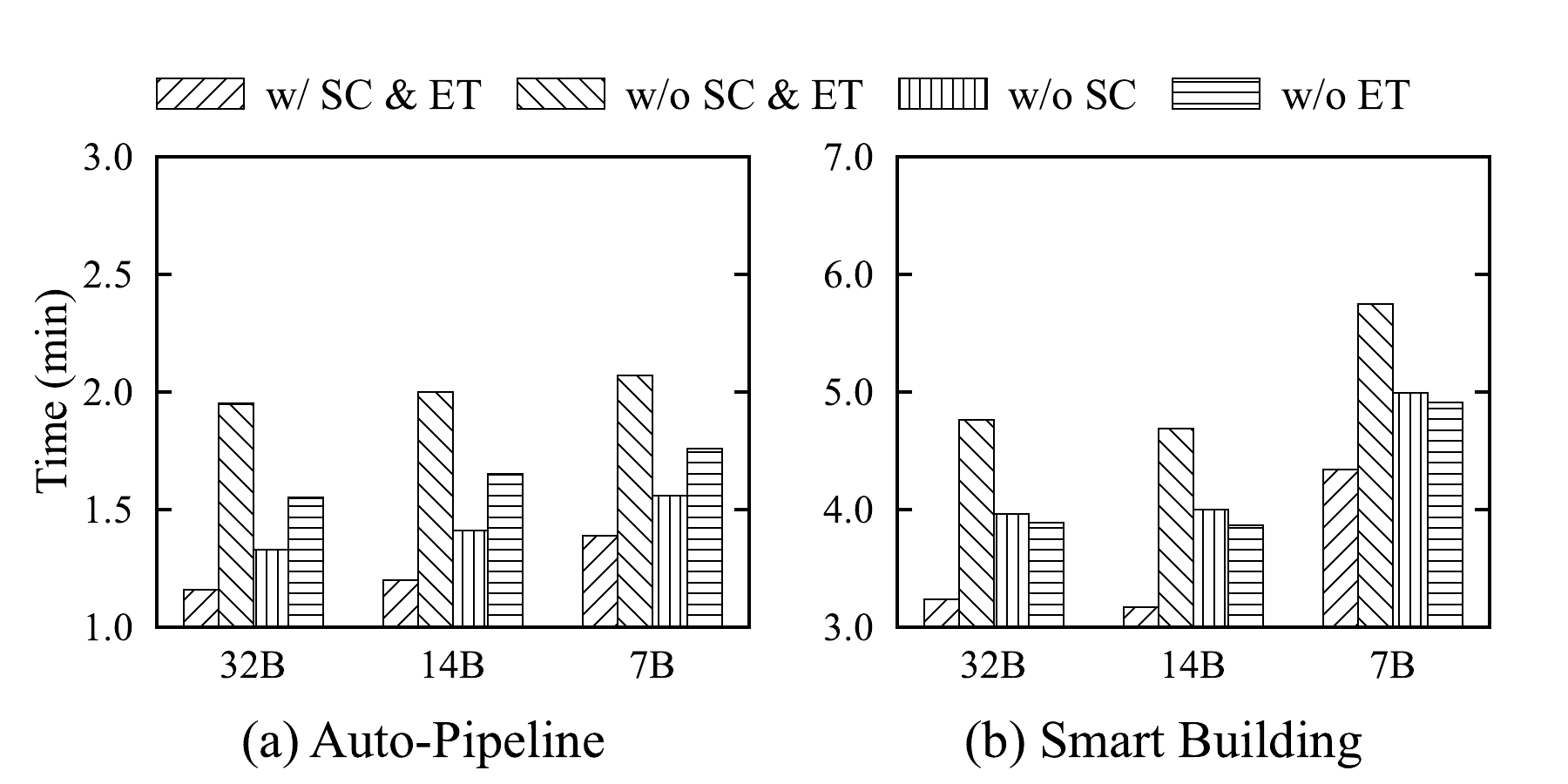}
  \vspace{-2em}
  \caption{Impact of simulation cache (SC) and early termination (ET) on runtime for different LLM sizes.}
  \label{fig:epo-ablation}
  \vspace{-1em}
\end{figure}

We conduct ablation studies on both datasets to evaluate the impact of the \emph{execution-aware pipeline optimizer} (\textsf{EPO}), which consists of two components: \emph{simulation cache} (SC) and \emph{early termination} (ET). 
We compare four configurations: enabling both SC and ET (denoted as w/ SC \& ET), disabling both (w/o SC \& ET), and disabling each individually (w/o SC and w/o ET, respectively).
Figure~\ref{fig:epo-ablation} reports the average runtime across different LLM sizes. Table~\ref{tab:epo-accuracy} shows the corresponding EX and CS. It is evaluated using Qwen-Coder-32B.

As shown in Figure~\ref{fig:epo-ablation}, removing either SC or ET results in increased runtime across all model sizes on both datasets. 
This verifies that both components contribute significantly to the efficiency of \name~. In particular, the absence of the simulation cache leads to redundant LLM queries during the MCTS search, as parameter predictions must be recomputed for the same node states.
Similarly, without early termination, the search continues exploring even after high-quality pipelines have already been found. Table~\ref{tab:epo-accuracy} shows that EX and CS vary slightly across different configurations.
Although there are some fluctuations, the variation in EX does not exceed 1\%, and the CS differences are minor.
This indicates that \textsf{EPO} improves efficiency without noticeably affecting overall performance. These results demonstrate that SC and ET are effective at reducing runtime without significantly impacting pipeline quality.

\vspace{-0.2em}
\subsection{Further analysis}
\label{further-analysis}

\noindent{\textbf{Effect of reward strategy.}}
To investigate the impact of reward strategy on pipeline quality and runtime, we compare three reward strategies: self reward, execution-aware reward, and hybrid reward, as summarized in Table~\ref{tab:reward_ablation}. Detailed information of the reward strategy can be found in Section~\ref{epo}. Several key observations can be made. (i) The execution-aware reward consistently outperforms the other two strategies in terms of both execution accuracy and column similarity. These results confirm that incorporating execution feedback provides a more reliable and precise evaluation of pipeline correctness. (ii) The hybrid reward strategy performs better than self reward. This suggests that combining LLM reasoning with execution feedback leads to more accurate pipeline evaluation. The feedback helps correct potential mistakes that the LLM might overlook when reasoning alone. (iii) Strategies that rely on large language models, such as hybrid reward and self reward, generally take a longer time to complete. This is because they involve additional inference steps during evaluation.

\begin{table}[t]
\centering
\caption{Effect of simulation cache and early termination on execution accuracy (EX) and column similarity (CS)}
\label{tab:epo-accuracy}
\vspace{-1em}
\setlength{\tabcolsep}{10pt}
\renewcommand{\arraystretch}{1.1}
\begin{tabular}{@{}l c@{\hskip 2pt}c c@{\hskip 2pt}c@{}}
\toprule
\multirow{2}{*}{\textbf{Configuration}} & \multicolumn{2}{c}{\textbf{Auto-Pipeline}} & \multicolumn{2}{c}{\textbf{Smart Building}} \\
\cmidrule(lr){2-3} \cmidrule(l){4-5}
 & EX(\%) & CS(\%) & EX(\%) & CS(\%) \\
\midrule
w/ SC \& ET     & 88.57 & 89.69 & 68.57 & 84.67 \\
w/o SC \& ET    & 86.69 & 87.55 & 69.38 & 82.39 \\
w/o SC          & 88.36 & 88.87 & 69.52 & 84.24 \\
w/o ET          & 87.44 & 89.54 & 68.57 & 85.39 \\
\bottomrule
\end{tabular}
\vspace{-0.5em}
\end{table}

\begin{figure}[t]
  \centering
  \vspace{-3mm}
  \includegraphics[width=\linewidth]{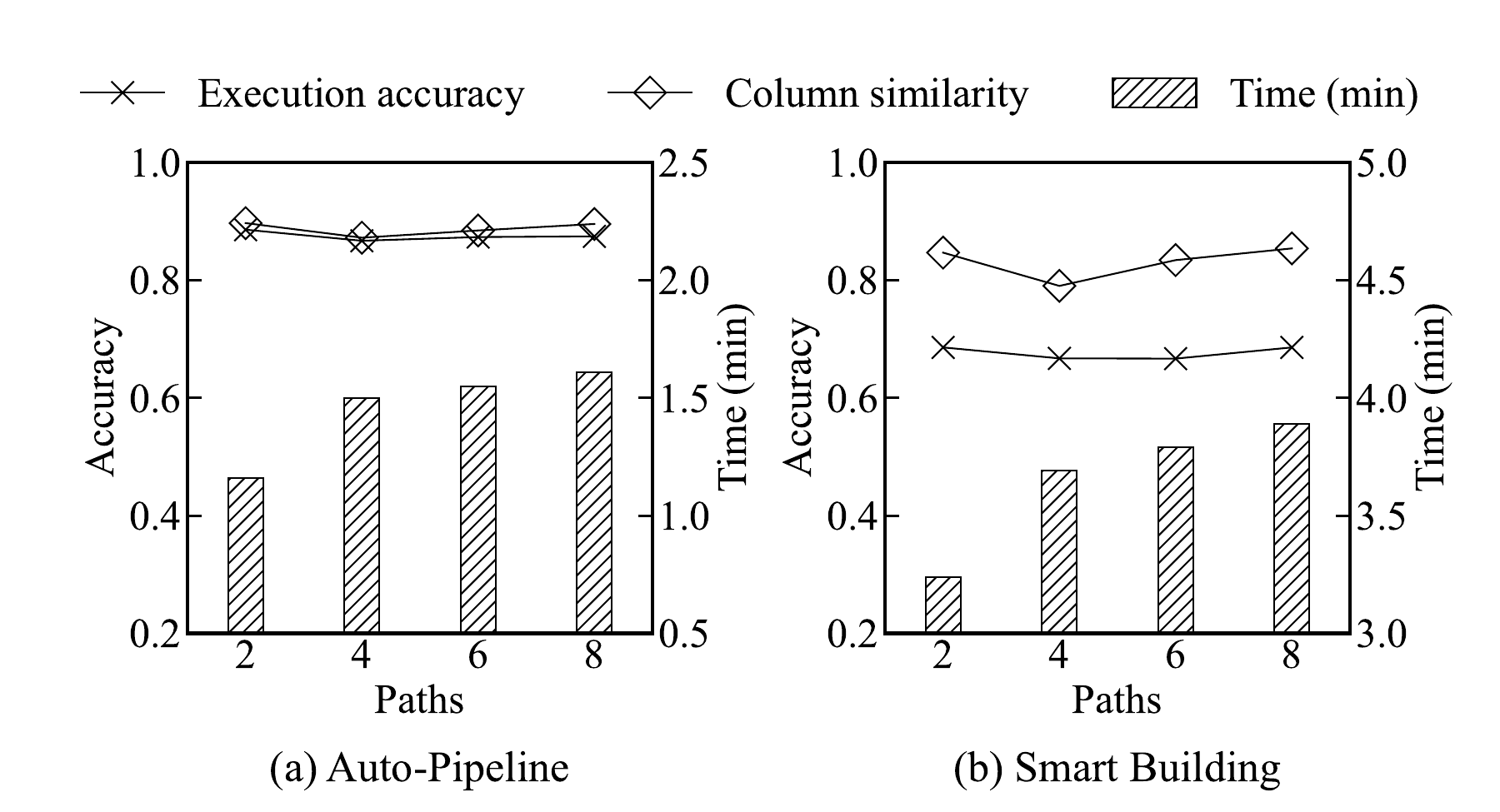}
  \vspace{-2em}
  \caption{Effect of exploration constant on ~\name.}
  \label{fig:terminal_early}
\end{figure}

\noindent{\textbf{Effect of early termination threshold.}}We study the impact of the early termination threshold $K$. As shown in Figure~\ref{fig:terminal_early}, increasing K first leads to a drop in performance, followed by a recovery. This trend is more pronounced on the Smart Building dataset, while Auto-Pipeline remains relatively stable. The initial drop in performance is likely due to the inclusion of lower-quality pipelines as the search explores more candidates. Eventually, performance recovers as more complete candidates are considered—but at a significant cost in runtime. Based on this observation, we set $K = 2$ as the default, as it achieves the best balance between accuracy and efficiency.



\begin{table}[t]
\caption{Effect of reward strategy on ~\name.}
\label{tab:reward_ablation}
\vspace*{-2.5mm}
\centering
\resizebox{\linewidth}{!}{
\begin{tabular}{l|l|ccc}
\toprule
\textbf{Benchmark} & \textbf{Reward strategy} & \textbf{EX(\%)} & \textbf{CS(\%)} & \textbf{Time (min)} \\
\midrule
\multirow{3}{*}{Auto-Pipeline} 
& Self reward       & 83.78 & 86.29 & 1.27 \\
& Execution-aware reward & \textbf{88.57} & \textbf{89.69} & 1.16 \\
& Hybrid reward        & 87.94 & 88.64 & 1.29 \\
\midrule
\multirow{3}{*}{Smart Building} 
& Self reward       & 44.76  & 54.34  & 3.49 \\
& Execution-aware reward & \textbf{68.57}  & \textbf{84.67}  & 3.24 \\
& Hybrid reward        & 59.05  & 69.67  & 3.54 \\
\bottomrule
\end{tabular}
}
\end{table}

\begin{figure}[t]
  \centering
  \vspace{-3mm}
  \includegraphics[width=0.98\linewidth]{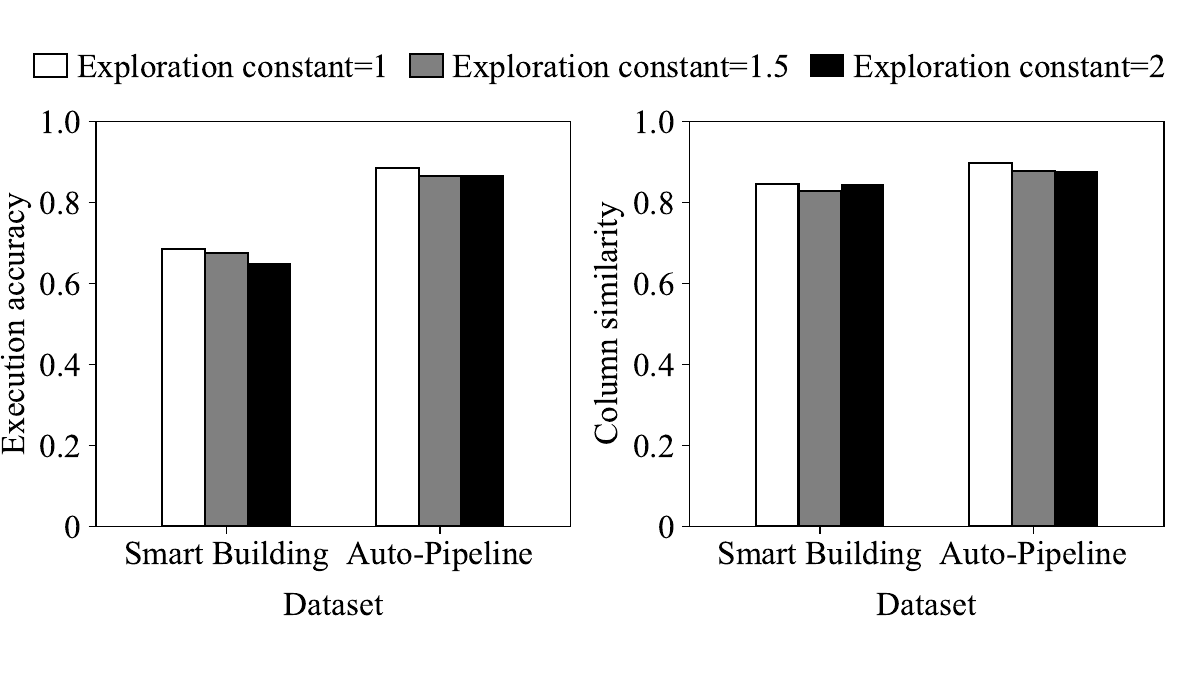}
  \vspace{-1em}
  \caption{Effect of exploration constant on ~\name.}
  \label{Exploration Constant}
  \vspace{-2em}
\end{figure}

\noindent{\textbf{Effect of exploration constant.}} We further investigate the effect of the exploration constant in MCTS by evaluating three settings: 1.0, 1.5, and 2.0. As shown in Figure~\ref{Exploration Constant}, minor fluctuations in execution accuracy and column similarity are observed as the constant increases, but the variation remains within a reasonable range. Overall, the constant does not significantly affect pipeline quality.

\section{Related Work}
\label{sec:related_work}

\noindent
\textbf{ADP methods.} Automatic Data Preparation (ADP)~\cite{autopipeline21, li2023auto, lai2025auto, tran2009query, bavishi2019autopandas} aims to empower non-technical users by automatically transforming raw data into a standard tabular format, ready for business intelligence (BI), data analytics, or machine learning. There is a large body of prior work on ADP. Transform-by-example (TBE) \cite{he2018transform, TDE18, DataXFormer16, gulwani2011automating, HeerHK15, Foofah17, jin2018clx, singh2016blinkfill, koehler2019incorporating} enables users to specify desired data transformations by providing one or more illustrative examples. Transform-by-pattern~\cite{jin2020auto} seeks to discover reusable transformation logic by analyzing large corpora of table columns and their structural patterns. Transform‑for‑joins~\cite{autojoin17, nobari2022efficiently} focuses on automatically generating equi-joins between tables with semantically matching, but textually differing, key columns. Transform-by-target~\cite{autosuggest20, autopipeline21} focuses on synthesizing end-to-end data pipelines that transform raw input tables into a predefined target table, leveraging functional dependencies and keys to guide the transformation process. The methods above focus on row-to-row transformations guided by partial specifications and rely on direct access to the target instances. This makes them orthogonal to the approach we explore in this work. Other methods that utilize target table instances for training, such as Auto-Tables~\cite{li2023auto} and Auto-Prep~\cite{lai2025auto}, are likewise orthogonal to our problem setting.

More recently, large language model (LLM), as a powerful tool, has facilitated the emergence of a new class of ADP methods, allowing data preparation without extensive supervision or access to target instances. Among LLM-based approaches, SQLMorpher~\cite{sharma2023automatic} tackles the ADP problem by prompting LLM to generate SQL transformation scripts. Its prompt incorporates attribute descriptions and schema change hints. It is the most comparable to our work in terms of both objective and formulation.
In addition, ChatPipe\cite{chen2024chatpipe} and Text-to-Pipeline\cite{ge2025text} explore a different direction—focusing on interactive, language-driven pipeline synthesis. These approaches rely on explicit user intent. They are different from our work.

\noindent
\textbf{LLM-based structured reasoning for ADP.} Recent advances also demonstrate that structured reasoning techniques can further enhance the reasoning capabilities of LLMs, especially in complex multi-step or zero-shot settings~\cite{qi2024mutual, chen2024alphamath, dainese2024generating, yao2023tree}. Chain-of-Thought (CoT) prompting~\cite{wei2022chain} enhances reasoning by generating intermediate natural language steps before producing transformation code. Chain-of-Table~\cite{wang2024chain} extends this by evolving intermediate table states, enabling LLMs to iteratively generate and apply operators, providing structured context throughout the transformation. ReAct~\cite{yao2023react} combines reasoning with action execution, incorporating feedback mechanisms like execution validation to improve the correctness of synthesized pipelines. FunctionCalling~\cite{qin2024tool} constrains LLMs to invoke a predefined set of structured transformation operators (e.g., GroupBy, Pivot, Join) with well-typed JSON arguments. Although these methods were not originally designed for ADP, they can be effectively adapted by modifying their reasoning procedures and aligning their operator sets with ADP tasks. Detailed adaptations are described in Section~\ref{baseline}.


\section{Conclusions}
\label{sec:conclusions}

In this paper, we present \name~, an open-source LLM powered automatic data preparation framework, which is driven by Monte Carlo Tree Search (MCTS).
In \name~, we first introduce a DP action sandbox (\textsf{DPAS}) to constrain the exploration space. Then, to navigate the complex problem of multi-step data preparation, we develop a fundamental pipeline generator (\textsf{FPG}).
It formulates pipeline synthesis as a MCTS problem guided by LLM-based action inference.
To further improve search reliability and efficiency, we design an execution-aware pipeline optimizer (\textsf{EPO}).
It incorporates execution-aware reward mechanisms to ensure the reliability and accelerates the search process via simulation caching and early termination.
Comprehensive experiments confirm the superiority of \name~.
In the future, a further direction is to incorporate user intent into the data preparation task.



\balance

\bibliographystyle{ACM-Reference-Format}
\bibliography{ref}

\end{document}